\documentclass{article} 
\usepackage{nips13submit_e,times}
\usepackage{hyperref}
\usepackage{url}
\usepackage[ruled]{algorithm}
\usepackage[noend]{algpseudocode}
\usepackage{amsmath}
\usepackage{amsthm}
\usepackage{amssymb}
\usepackage{graphicx} 
\usepackage{comment}
\newtheorem{thm}{Theorem}

\newcommand{\R}{\mathbb{R}}

\newcommand{\N}{\mathcal{N}}
\newcommand{\mat}[1]{{\mathbf{#1}}}

\title{Combining Structured and Unstructured Randomness in 
Large Scale PCA}

\author{
Nikos Karampatziakis\\
Microsoft \\
1 Microsoft Way \\
Redmond, WA 98052 \\
\texttt{nikosk@microsoft.com} \\
\And
Paul Mineiro \\
Microsoft \\
1 Microsoft Way \\
Redmond, WA 98052 \\
\texttt{pmineiro@microsoft.com} \\
}

\nipsfinalcopy 

\begin{document}

\maketitle

\begin{abstract}
Principal Component Analysis (PCA) is a ubiquitous tool with many
applications in machine learning including feature construction, subspace
embedding, and outlier detection.  In this paper, we present an algorithm
for computing the top principal components of a dataset with a
large number of rows (examples) and columns (features).   
Our algorithm leverages both structured and unstructured random
projections to retain good accuracy while being computationally efficient.
We demonstrate the technique on the winning submission the KDD 2010 Cup.
\end{abstract}

\section{Introduction}

PCA~\cite{Pearson1901}, and the related Singular Value Decomposition
(SVD), are versatile tools in machine learning because they can
compress a high dimensional dataset to a small number of dimensions. This
compression can uncover hidden patterns in the data~\cite{eigenfaces},
reduce noise, or facilitate the application of algorithms that need
not scale well to high dimensional data.  In a nutshell, if 
$\mat{X} \in \R^{n\times p}$ is the data matrix having $n$ examples and $p$ features,
whose SVD is $\mat{X} = \mat{U} \mat{\Sigma}\mat{V}^\top$, then
$\mat{U}\mat{\Sigma}$ are the \emph{principal components} and $\mat{V}$ is 
called the \emph{loadings}. Furthermore, computing the principal components 
of a new example $x$ can be done via $\mat{V}^\top x$. 
Classic algorithms for PCA
cannot scale to large datasets which has led to the recent interest in
randomized SVD algorithms~\cite{halko2011finding,redsvd,gensim}.
When the data is nominally $p$-dimensional but distributed close to a
$k$-dimensional subspace, these algorithms just need two streaming passes
over the data plus an additional $O(pk^2)$ processing to compute
the top $k$ singular values and vectors with very good accuracy. In
addition they require $O(pk)$ memory as working storage, and in a
distributed context require $O(pk)$ communication between worker nodes.

As the volumes of datasets increase, the limits of the randomized
algorithms are being stretched. Datasets with millions of examples and
millions of dimensions are common nowadays and by industrial standards
they are considered small. As an example, when $p=10^8$ and $k=300$, the
memory requirements of randomized algorithms are about 223 GB.  However,
existing randomized algorithms reveal the actual top loadings and principal components,
whereas  in machine learning we typically care only about the \emph{mapping}
$\R^p \to \R^k$ that projects a high dimensional vector
$x$ into the top principal component space.  In this paper,
we propose an algorithm to efficiently and accurately approximate this
mapping with reduced storage requirements. We do this by first 
applying a structured random projection to the data vectors 
to a lower dimension $d \ll p$ and then relying on the existing
randomized algorithms for truncated SVD.   With high probability, our 
resulting projection is onto a subspace very close to the top
principal component subspace. Our implementation interleaves the 
application of structured random projection and the steps of 
randomized SVD which leads to minimal overhead.  In this manner, 
both working storage and communication requirements are
reduced from $O(pk)$ to $O(dk)$.

Herein we focus on 
hashing as our way of computing structured random projections.
Hashing preserves sparsity, which fits well with the 
sparse datasets we use in the experiments as well as other
modern datasets arising from text and social graphs.
Other structured random projections such as subsampled fast
transforms (e.g.\ Hadamard, Hartley) can  
be used, and should be used for dense data.  

A natural question is whether both the structured randomness
we employ and the unstructured randomness\footnote{Randomized SVD implementations can use structured randomness
    instead of a Gaussian random projection.  However, the variants using
    structured randomness are slightly less accurate.}
already existent in randomized SVD algorithms are necessary. 
Couldn't we just project upfront down to $O(k)$ dimensions?
However, projecting directly to $O(k)$
dimensions loses a lot of accuracy. In this sense our algorithm interpolates
between two extremes: a fast but crude upfront (structured) random projection
followed by PCA and a slow but accurate randomized SVD algorithm that requires
orthogonalization of a $p\times k$ matrix.  In practice, randomized SVD
is constrained by the space to store, and time to orthogonalize, a
$p\times k$ matrix. We identify a parameter $d$, the size of upfront
random projection, that should match the available hardware, while causing 
low distortion. For a commodity machine $d=10^5$ to $d=10^6$ is
typical.

\subsection{Relation to Prior Work}

Randomized algorithms for numerical linear algebra have recently gained much
attention in theoretical computer science~\cite{Mahoney}. In this work we are
particularly interested in fast algorithms for truncated SVD such as those
developed in Halko et.~al.~\cite{halko2011finding}. Using randomization to speed
up SVD goes back at least a decade~\cite{Papadimitriou}, when Papadimitriou
et.~al.~made similar arguments to ours regarding the impact of random
projections without orthogonalization on SVD.  More recent advances have led to
very sharp bounds~\cite{halko2011finding}, and practically useful
algorithms~\cite{gensim,redsvd,rokhlin2009randomized,halko2011algorithm} at
least for datasets with either not too many rows or not too many columns.
Unfortunately in machine learning we are often faced with datasets where both
the number of examples and the number of features exceed the limits of the
randomized SVD algorithms in today's hardware. The use of structured randomness
allows us to reduce the number of features to a size that the randomized SVD
algorithms can handle, without distorting the final embedding too much. The
single hash utilized here is a computationally convenient technique.  There
exist more complicated hashing-based dimensionality reduction techniques with
superior inner product preservation guarantees~\cite{Dasgupta2010} for which
analogous arguments hold.  

\section{The Algorithm}

Let $\mat{X} \in \R^{n\times p}$ be the data matrix. We assume that the features
have zero empirical mean.\footnote{Uncentered data requires a rank-one
modification to Algorithm~\ref{alg:truncatedpca}, requiring an additional $O
(d)$ space.}  The principal components can be computed via the SVD $\mat{X} =
\mat{U} \mat{\Sigma}\mat{V}^\top$ where $\mat{U} \in \R^{n\times n}$ and
$\mat{V} \in \R^{p \times p}$ are orthogonal matrices and $\mat{\Sigma} \in
\R^{n \times p}$ is a (rectangular) diagonal matrix with
$\mat{\Sigma}_{ii}$ arranged in non-ascending order.  Truncating by
retaining the top $k$ singular values and corresponding vectors yields the best
(in Frobenius norm~\cite{eckart1936approximation}) rank-$k$ approximation of
$\mat{X}$: $\mat{\tilde{X}}=\mat{U}_{k} \mat{\Sigma}_{k} \mat{V}_{k}^\top$,
where $\mat{U}_k \in \R^{n\times k}$, $\mat{\Sigma}_k \in \R^{k \times k}$, and
$\mat{V}_k \in \R^{p \times k}$.  The \emph{whitened} PCA projection of a new
example is given by $\mat{\Sigma}_k^\dagger \mat{V}_k^\top x$, where
$\mat{\Sigma}^\dagger$ indicates the Moore-Penrose pseudo-inverse of
$\mat{\Sigma}$,  and Theorem~\ref{thm:subspace} will show that this mapping can
be approximated by our algorithm.

Obtaining $\mat{V}_k$ and $\mat{\Sigma}_k$ could be done with any SVD algorithm, 
however, once $n$ or $p$ is large, only the randomized SVD algorithms are practical.
The randomized algorithms work in two phases. In the first phase they 
probe the range of the input matrix with a random matrix $\mat{\Omega}$.
They potentially perform multiple passes over the data, though here we will only assume one pass.
Next, they orthogonalize the image of $\mat{\Omega}$ under the input matrix and project onto that
basis in the second pass. Even though these algorithms have been previously adapted for 
PCA~\cite{rokhlin2009randomized,halko2011algorithm}, they 
assume that the orthogonalization step can be done efficiently. 
This is only true if either $n$ or $p$ is not too large, but not both.

We can easily eliminate the dependence on $n$ by looking at the empirical covariance
matrix $\frac1n \mat{X}^\top \mat{X}$, whose top eigenvectors are $\mat{V}_k$. 
We can then apply a randomized SVD algorithm on this matrix.
A two-pass randomized algorithm with
orthogonalization of columns has space complexity $O(p k)$ and time complexity
$O (p k^2)$; furthermore computing the image of $\mat{\Omega}$ under the
empirical covariance is data parallel as can be seen by
$ 
   \frac1n \mat{X}^\top \mat{X} \mat{\Omega} = \frac 1n \sum_{i=1}^n x_i x_i^\top \mat{\Omega},
$
where $x_i$ is the $i$-th example.  This procedure produces  
$\mat{V}_k$ and $\mat{\Sigma}_k$ which can be used in a
subsequent pass over the data to produce the (whitened) principal components. 
In practice, if the orthonormal basis for the column (feature) space fits into main memory, the
algorithm is very fast, and so is suitable for large data sets where the number
of features is modest, up to circa $p = 10^6$ on current commodity hardware.

\begin{algorithm}[t]
    \caption{HPCA: Truncated PCA via Hashing/Hadamard/Hartley}
  \label{alg:truncatedpca} 
  \begin{algorithmic}[1]
    \Function{HPCA}{$\mat{X}_{n \times p}, \mat{H}_{p \times d}, k, d$} \Comment{$\mat{H}$ not materialized}
      \State $\mat{\Omega}_{d \times k} \leftarrow$ Random Gaussian matrix ($\Omega_{ij} \sim \N (0, 1)$)
      \State $\mat{Y}_{d \times k} \leftarrow (\mat{X}\mat{H})^\top (\mat{X}\mat{H}) ~\mat{\Omega} ~/~ n$ \label{firstpass} \Comment{First data-parallel pass}
      \State $\mat{Q}_{d \times k} \leftarrow$ Gram-Schmidt orthonormalization of the column space of $\mat{Y}$.\Comment{$O (d k)$ space}
      \State $\mat{Z}_{d \times k} \leftarrow (\mat{X}\mat{H})^\top (\mat{X}\mat{H}) ~\mat{Q} ~/~ n$ \label{secondpass} \Comment{Second data-parallel pass}
     \State $\mat{\Upsilon} \mat{\tilde{\Sigma}_1}^4 \mat{\Upsilon}^\top \leftarrow$ Spectral decomposition of $\mat{Z}^\top \mat{Z}$
     \Comment{$\mat{Z}^\top \mat{Z} \in \R^{k \times k}$} \label{transpose}
     \State \Return{($\mat{\tilde{V}_1} = \mat{Z} \mat{\Upsilon}(\mat{\tilde{\Sigma}_1}^2)^\dagger $, $\mat{\tilde{\Sigma}_1}$)} \Comment{$\Sigma ^\dagger$ is the Moore-Penrose pseudo-inverse of $\Sigma$}
    \EndFunction
  \end{algorithmic}
\end{algorithm}

For datasets with hundreds of millions of features, such as the adjacency matrix
of an online social network, the space complexity associated with
orthogonalizing the approximate basis is impractical.  We therefore propose
using structured randomness, without explicit materialization of the projection
matrix and without orthogonalization, to reduce the number of features into the
zone where randomized SVD algorithms are viable.  Although not materialized, we
represent the structured randomness as a matrix $\mat{H} \in  \R^{p \times d}$.
By interleaving the structured randomness with the randomized algorithm, we
arrive at Algorithm~\ref{alg:truncatedpca}. This mainly follows the algorithm
from~\cite{halko2011finding} except at step~\ref{transpose} where we use that
the covariance is symmetric to write it as in line~\ref{firstpass}.  We also
compute the spectral decomposition of a $k \times k$ matrix instead of the SVD
of a $d \times k$ matrix as it is sufficient to extract the loadings and the
singular values.

For sparse data, e.g.\ text or social graphs, we find a hash based structured randomness~\cite{Weinberger2009} to
be computationally convenient and empirically effective.  Conceptually,
this scheme multiplies the data by a hashing matrix $\mat{H} \in \R^{p \times d}$ which is determined by
two hash functions $h: \{ 1, \ldots, p \} \to \{ 1,\ldots, d \}$ and $\xi: \{ 1, \ldots, p \} \to \{ \pm 1 \}$,
with $H_{ij}|_{\xi,h} = \xi(i) 1_{h(i) = j}$.  For analytical purposes, the hash functions $h$ and $\xi$ are 
considered drawn uniformly at random from a universal family. For dense data
(not presented here), subsampled Hadamard or Hartley transforms can be used.

\subsection{Analysis}\label{subsec:analysis}

Our proposed algorithm is a composition of dimensionality reduction via
structured randomness and established randomized SVD techniques.  Because the error
properties of randomized SVD techniques are well understood, we will focus
on the impact of dimensionality reduction on an exact truncated SVD.
Let $\mat{X} \mat{H} = \mat{\tilde U} \mat{\tilde \Sigma} \mat{\tilde
V}^\top$ be the SVD of the projected version of the data matrix, and let
$\mat{\tilde U_1}$, $\mat{\tilde \Sigma_1}$, and $\mat{\tilde V_1}$
be the truncated SVD components as follows
\begin{equation}\label{eqn:truncatedSVD}
\mat{X} \mat{H} = \bordermatrix{ ~ & k & n - k \cr ~ & \mat{\tilde U_1} & \mat{\tilde U_2} } \bordermatrix{ ~ & k & n - k \cr ~ & \mat{\tilde \Sigma_1} & \mat{0} \cr ~ & \mat{0} & \mat{\tilde \Sigma_2}}  \bordermatrix{ ~ & d \cr ~ & \mat{\tilde V_1}^\top \cr ~ & \mat{\tilde V_2}^\top \cr} \begin{array}{c} k \\  n - k \end{array},
\end{equation}
where $k < \min\{ d, n \}$.  Although our algorithm manipulates a
transformed version of the empirical covariance matrix $\mat{H}^\top
\mat{X}^\top \mat{X} \mat{H}$, it is more convenient to consider
the transformed version of the Gram matrix $\mat{X} \mat{H}
\mat{H}^\top \mat{X}^\top$. We do this because the Gram matrix has the 
same nonzero eigenvalues as the covariance matrix and the corresponding 
eigenvectors are the whitened principal components.  In particular, any mapping $\mat{H}$ which
approximately preserves inner products will lead to a Gram matrix that 
is close to the original one in Frobenius norm.  In particular, hash based structured 
randomness approximately preserves inner products~\cite{Weinberger2009}.

Since the perturbation of the Gram matrix has small Frobenius norm,
we can apply a classic theorem by Wedin~\cite{Wedin1972}, which states
that the left singular subspaces associated with the original and perturbed 
matrices will be close (in the sense of a small canonical
angle) if two things are true: first, if there is a gap between the
$k^\mathrm{th}$ largest singular value of the perturbed matrix and the
$(k+1)^{\mathrm{th}}$ largest singular value of the original matrix;
and second, if the difference between the Gram matrices has small norm.  This reasoning leads
to the following theorem.
\begin{thm}[Subspace Approximation]\label{thm:subspace} 
Let $\mat{X}$ be a data matrix with $n$ rows, and let $\mat{\tilde{X}} =
\mat{X} \mat{H}$ be the hashed data matrix.  Let $\mat{X} \mat{X}^\top$
and $\mat{\tilde{X}} \mat{\tilde{X}}^\top$ have spectral
decompositions $\mat{U} \mat{\Sigma}^2 \mat{U}^\top$ and $\mat{\tilde{U}}
\mat{\tilde{\Sigma}}^2 \mat{\tilde{U}}^\top$ respectively, conformally
partitioned as in \eqref{eqn:truncatedSVD}. Let $\mat{\Phi}$ be the matrix
of canonical angles between the column spaces $\mathcal{R} (\mat{U_1})$ of
$\mat{U_1}$ and $\mathcal{R} (\mat{\tilde{U}_1})$ of $\mat{\tilde{U}_1}$.  
Suppose that there are numbers $\alpha, \gamma > 0$ such that
\begin{displaymath}
\begin{aligned}
\min_{\tilde{\mu} \in \sigma (\mat{\tilde{\Sigma}}_1^2)} \tilde{\mu} \geq \alpha
+ \gamma & \;\mathrm{and}\; & \max_{\mu \in \sigma (\mat{\Sigma}_2^2)} \mu \leq
\alpha,
\end{aligned}
\end{displaymath}
where $\sigma (\mat{\Sigma})$ denotes the set of diagonal values.
Let $\mathop{r} (\mat{X})$ denote the set of rows of $\mat{X}$. Define
\begin{gather*}
\eta = \max_{x, x' \in \mathop{r}(\mat{X})} \left( \frac{ \| x \|_{\infty} }{ \| x \|_2 }, \frac{ \| x' \|_{\infty} }{ \| x' \|_2 }, \frac{ \| x - x' \|_{\infty} }{ \| x -  x'\|_2 } \right).
\end{gather*}
If $d \geq  144 \log (n / \delta) / \epsilon^2$ and $\eta \leq \epsilon /
(18 \sqrt{2 \log (n / \delta) \log (d / \delta)})$, then with probability
at least $6 \delta$ with respect to the uniform distribution over
functions $h: \{ 1, \ldots, p \} \to \{ 1, \ldots, d \}$ and $\xi: \{
1, \ldots, p \} \to \{ \pm 1 \}$,
\begin{displaymath}
\left\| \sin \mat{\Phi} \right\|_F \leq O \left(\frac{\epsilon}{\gamma} \right).
\end{displaymath}
\end{thm}
Theorem~\ref{thm:subspace} indicates that under appropriate
conditions, the top whitened principal components induced by the hashed projection
$\mat{\tilde{U}_1}^\top = \mat{\tilde{\Sigma}_1}^\dagger \mat{\tilde{V}_1}^\top
\mat{H}^\top \mat{X}^\top$ will be in a space close to the whitened principal
components obtained from the exact projection $\mat{U_1}^\top = \mat{\Sigma}_1^\dagger
\mat{V_1}^\top \mat{X}^\top$.  An important condition is a gap between the
$k^\mathrm{th}$ original singular value and the $(k+1)^\mathrm{th}$
perturbed singular value. As can be seen in the full proof, the Frobenius norm of the
difference between the Gram matrices need only be bounded with high probability once. 
Then Wedin's deterministic error bound can be applied for all $k$.  Therefore,
if there is a large spectral gap at any $k$ and we truncate the
PCA at $r \geq k$, then the subspaces induced by the top $k$ whitened components
will be close.

\section{Experiment}

The winning submission to the 2010 KDD Cup was a linear model developed using
extensive feature engineering~\cite{yu2010feature}.  We focus on the winning design matrix because it 
is publicly available, large enough to prohibit exact 
decomposition, but small enough to admit standard two-pass randomized decomposition.  Specifically we used the \texttt{kdda} dataset\footnote{http://www.csie.ntu.edu.tw/~cjlin/libsvmtools/datasets/binary.html\#kdd2010 (algebra)}
which consists of $8.4$ million training examples, $20.2$ million 
features, and 314 million non-zero entries.

We used REDSVD~\cite{redsvd} in sparse PCA mode as a baseline, which implements a 
two-pass randomized decomposition with a materialized Gaussian matrix.  Computing the top
40 principal components consumes 17.5 gigabytes of RAM and takes 1591 seconds to 
compute the loadings and factor scores.  On the same machine, algorithm~\ref{alg:truncatedpca} with
$d=10^6$ uses 780 megabytes of RAM and takes 1030 seconds to compute the loadings and
factor scores.  The decrease in memory usage is expected (due to $p/d \approx 20$), while the 
decrease in computation time is mostly attributable to the orthogonalization step being faster 
due to working with smaller vectors.

To motivate the use of PCA in this setting, we augmented the design matrix by
interacting the raw feature values  with the first few principal components 
as computed by HPCA, but
otherwise did not alter the baseline training procedure of the winning
submission.  With only 5 components the improvement in performance
exceeds the difference between the winning and $2^\mathrm{nd}$ place score.

\section{Conclusion}

In this paper we analyze theoretically and empirically the composition
of structured and unstructured randomness with established randomized SVD techniques for the purpose
of computing PCA components.  Theoretically we leverage inner-product
preservation guarantee to show the resulting PCA components are close to those computed
by fully unstructured randomness.  Empirically, the resulting algorithm
is so highly scalable that experiments with some of the largest publicly
available matrices were easily implemented on a commodity laptop.  

\subsubsection*{References}


\begingroup
\renewcommand{\section}[2]{}%
\bibliography{svdpaper}{}
\endgroup
\bibliographystyle{plain}


%
%


\end{document}